\documentclass[conference]{IEEEtran}
\IEEEoverridecommandlockouts
\usepackage{cite}
\usepackage{amsmath,amssymb,amsfonts}
\usepackage{algorithmic}
\usepackage{graphicx}
\usepackage{amsmath,amssymb}
\usepackage{booktabs}      
\usepackage{multirow}       
\usepackage[table]{xcolor}  
\usepackage{adjustbox}
\usepackage[ruled,vlined]{algorithm2e}
\usepackage{colortbl}       
\definecolor{Gray}{gray}{0.9} 
\usepackage{textcomp}
\usepackage{xcolor}
\def\BibTeX{{\rm B\kern-.05em{\sc i\kern-.025em b}\kern-.08em
    T\kern-.1667em\lower.7ex\hbox{E}\kern-.125emX}}
\begin{document}

\title{DPDEdit: Detail-Preserved Diffusion Models for \\ Multimodal Fashion Image Editing}

\author{
Xiaolong Wang\textsuperscript{1}, 
Zhiqi Cheng\textsuperscript{2}, 
Jue Wang\textsuperscript{3}, 
Huizi Xue\textsuperscript{1}, 
Xiaojiang Peng\textsuperscript{1}\textsuperscript{*}%
\thanks{*Corresponding author: pengxiaojiang@sztu.edu.cn} \\
\textsuperscript{1}Shenzhen Technology University, China \\
\textsuperscript{2}University of Washington, USA \\
\textsuperscript{3}Shenzhen Institute of Advanced Technology, Chinese Academy of Sciences, China
}

\maketitle

\begin{abstract}
Fashion image editing is a crucial tool for designers to convey their creative ideas by visualizing design concepts interactively. However, current fashion image editing techniques often struggle to accurately identify editing regions and preserve the desired garment texture detail. To address these challenges, we present Detail-Preserved Diffusion Models (DPDEdit), a new multimodal fashion image editing architecture based on latent diffusion models. 
To precisely locate the editing region, we introduce Grounded-SAM to predict the editing region. To transfer the detail of the given garment texture into the target image, we propose a texture injection and refinement mechanism. This mechanism employs a decoupled cross-attention layer to integrate textual descriptions and texture images, and incorporates an auxiliary U-Net to preserve the high-frequency details of generated garment texture.
Additionally, we extend the VITON-HD dataset using a multimodal large language model to generate paired samples with texture images and textual descriptions. Extensive experiments show that our DPDEdit outperforms state-of-the-art methods in terms of image fidelity and coherence with the given multimodal input.
\end{abstract}



\section{Introduction}
The purpose of fashion image editing is to manipulate fashion images according to the user's creative vision, thereby materializing their fashion concepts. 
This approach provides a seamless interface for both designers and non-experts to explore and visualize their fashion ideas. Furthermore, fashion image editing algorithms hold significant promise for e-commerce, advertising, and social networks. As computer vision increasingly intersects with the fashion industry\cite{tryon1,tryon2,clothrecommend2}, there is growing research interest in this emerging field\cite{pernuvs2023fice,MGD,texfit}.

Previous works\cite{zhu2017your,jiang2022text2human,pernuvs2023fice} has attempted to use GAN-based methods to generate and edit fashion images based on textual descriptions. Although GANs have shown potential, they are often plagued by issues related to training instability and struggle to produce high-quality generated images with abundant details. In contrast, Diffusion Models\cite{diffusion1,diffusion2,stablediffusion} have emerged as a promising alternative for image editing tasks, recognized for their ability to produce high-quality results and provide more stable and controllable generation mechanisms. TexFit\cite{texfit} introduces a straightforward text-driven fashion image editing method based on diffusion models.It is user-friendly and generates impressive results. However, relying exclusively on textual input poses challenges in accurately capturing the user's design specifications, including garment styles, patterns, and fabric textures. This limitation often results in discrepancies between the generated images and the user's intended vision.
\begin{figure}[t]
    \centering
    \includegraphics[width=1.0\linewidth]{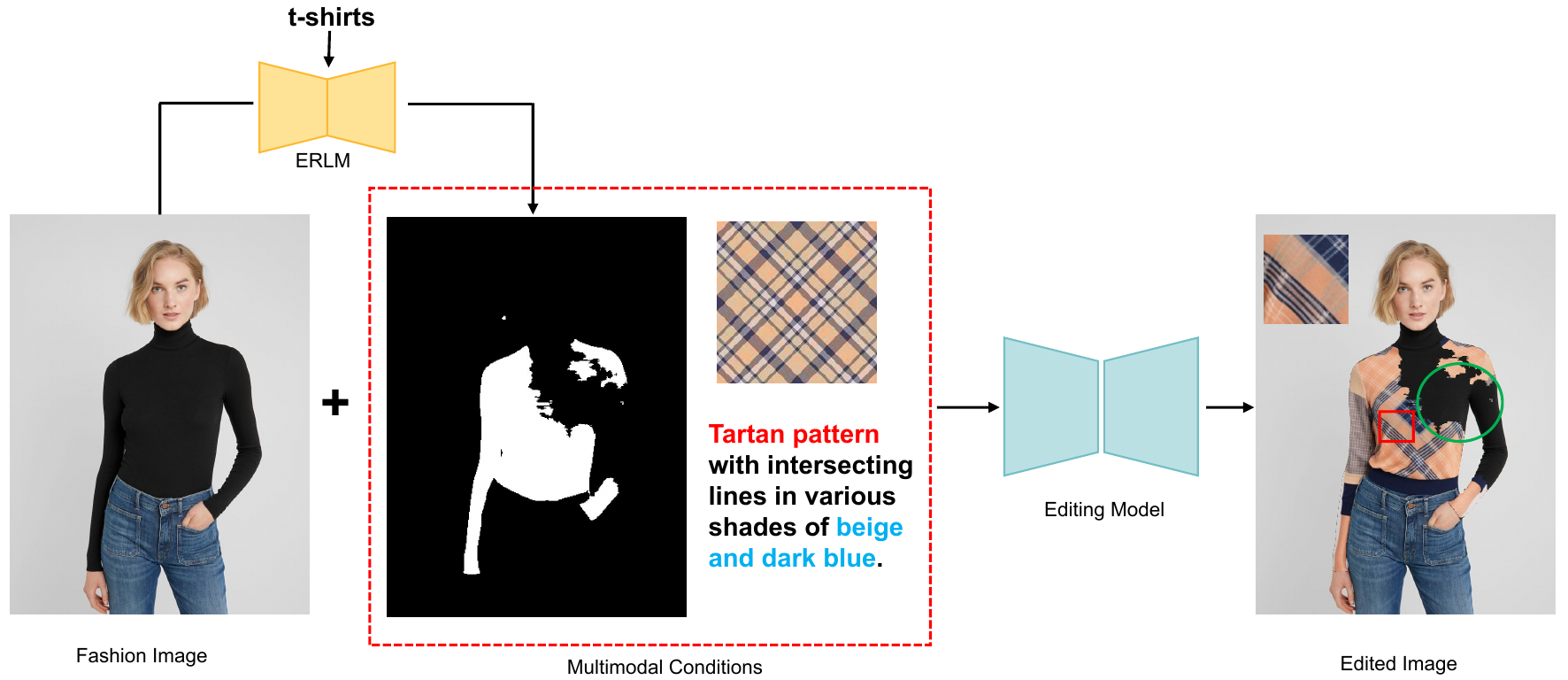}
    \caption{\small Drawbacks of the existing fashion editing pipeline, specifically in accurately identifying the editing regions (green region) and in maintaining consistency in the garment texture (red region).}
    \label{fig:0}
    \vspace{-0.5em}
\end{figure}

\begin{figure*}[t]
    \centering
    \includegraphics[width=0.9\textwidth]{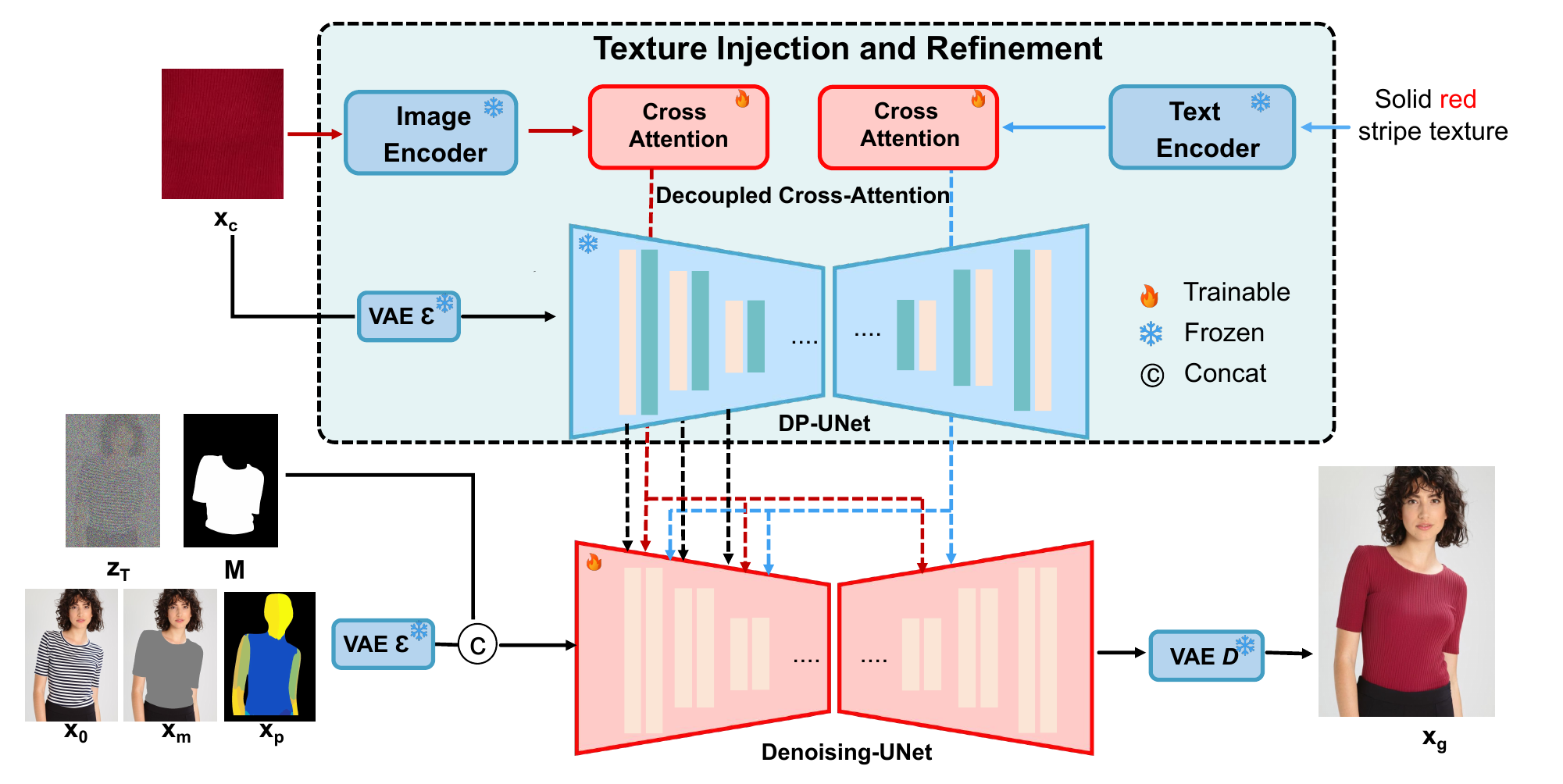}
    \caption{\small Overview pipeline of DPDEdit. The inputs of Denoising-UNet include the noisy latents \( z_T \) derived from the latent representation \( \mathcal{E}(x_0) \), along with the inpainting mask \( M \), masked image \( \mathcal{E}(x_m) \), and DensePose image \( \mathcal{E}(x_p) \). The fire icon indicates that the module's parameters require tuning, while the snowflake icon denotes modules that do not require tuning. }
    \label{fig:3}
\end{figure*}

As a result, introducing  multimodal approaches in
fashion image editing is essential for meeting
user requirements.IP-Adapter\cite{ipadapter} introduces a lightweight adapter that injects image conditions into the denoising process.  MGD\cite{MGD} integrates text, human pose, and garment sketch modalities for fashion image editing using text inversion techniques. Ti-MGD\cite{Ti-MGD} further incorporates clothing texture control. Although Ti-MGD incorporates multimodal conditional control to generate garment texture information, relying exclusively on CLIP\cite{clip} for extracting texture image features hinders the accurate restoration of complex and detailed textures. Additionally, these methods lack an emphasis on the precise localization of the editing region, limiting their effectiveness as a general-purpose solution. TexFit proposes a Editing Region Location Module (ERLM), which generates corresponding editing region masks using an encoder-decoder architecture. However, we found that this method falls short when dealing with fashion images involving complex human poses and diverse clothing styles. These limitations are illustrated in Figure \ref{fig:0}.

To address the aforementioned drawbacks, we introduce Detail-Preserved Diffusion Edit(DPDEdit) method, which integrates multiple modalities within a latent diffusion model for fashion image editing. DPDEdit leverages multimodal inputs, including text, human densepose\cite{densepose}, region mask and texture images to guide the garment editing process. To locating editing regions in complex scenarios, we utilize the latest research advancement, Grounded-SAM, for garment region segmentation. Grounded-SAM leverages its powerful segmentation capabilities to accurately generate a mask for the editing region based on the user's text prompt. In order to align the generated garment with the input texture image, we propose a texture injection and refinement mechanism. This mechanism employ a decoupled cross-attention layer to effectively guide the diffusion process under the joint control of texture image and textual description.To preserve intricate garment textures and enhance fine details, we employed a pre-trained auxiliary U-Net, named Detail-Preserved U-Net(DP-UNet), to extract high-frequency features from the texture images and integrate them into the denoising-UNet. DP-UNet supplements the texture image details, ensuring that the generated garments closely align with the input texture patterns.

To the best of our knowledge, there is no publicly available dataset that includes both garment texture images and corresponding text descriptions. To address this gap and meet the requirements of our task, we have extended the VITON-HD dataset \cite{viton-hd}. Specifically, we extracted fabric texture images from the garment images in the original dataset. Using the Multimodal Large Language Model LLaVA \cite{llava}, we generated appropriate captions for these fabric texture images, thereby creating a paired text-image dataset suitable for training and evaluation.

In summary, our contributions are threefold: (1) We propose the DPDEdit framework for fashion image editing, which leverages multimodal inputs to guide the diffusion model. This approach generates high-quality images that are consistent with the input modalities and allows for fine-grained control over the fabric texture of the clothing. (2) We employ Grounded-SAM to accurately identify the editing region and introduce a texture injection and refinement mechanism to preserve the intricate details of the garment texture, aligning with the specific requirements of our task. (3) We have extended the VITON-HD dataset to include fabric texture images of garments along with corresponding text captions, providing a valuable resource for future research in this domain.

\begin{figure}[t]
    \centering
    \includegraphics[width=0.9\columnwidth]{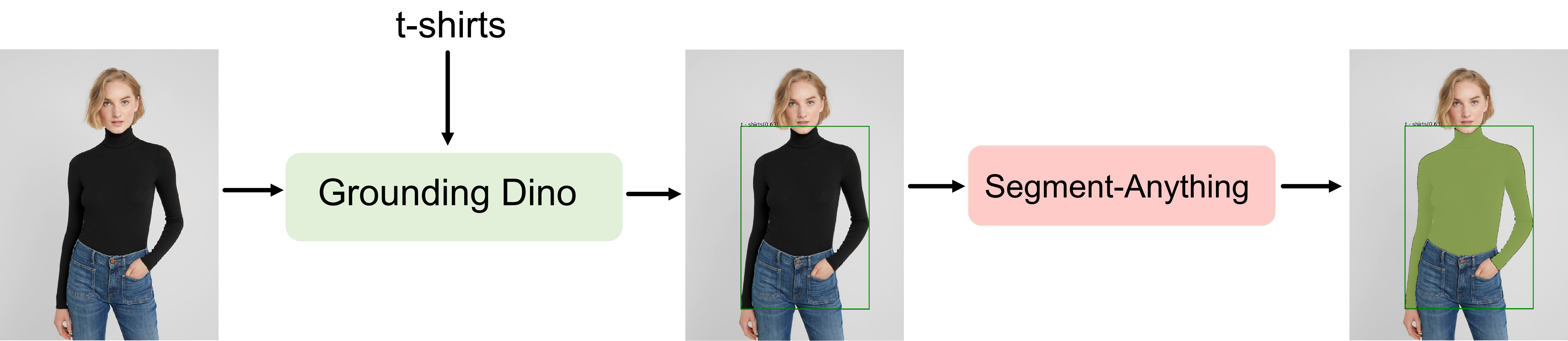}
    \caption{\small Illustration of the Grounded-SAM workflow}
    \label{fig:2}
    \vspace{-1.0em}
\end{figure}

\section{Methodology}
In this section, we propose a novel task to automatically edit fashion images conditioned on multiple modalities. Specifically, given the model image \( x_0 \), the name of the garment to be edited \( Y_0 \), Densepose \( x_{\text{p}} \) of the model image, fabric texture image \( x_c \), and the corresponding caption \( s \), we aim to generate a new image \( {x}_g \) that retains the information of the input model while replacing the target garment according to the multimodal inputs. An overview of our model is illustrated in Figure \ref{fig:3}.

\subsection{DPDEdit Framework}
We introduce the DPDEdit framework, which integrates Grounded-SAM for precise localization of editing regions and a main denoising U-Net for image generation.

\subsubsection{Grounded-SAM}
To achieve high-quality fashion image editing, precise identification and segmentation of the editing regions are essential. We utilize \textit{Grounded-SAM}, integrating \textit{Grounding-DINO} \cite{dino} and \textit{SAM} (Segment Anything Model) \cite{sam}, to ensure accurate localization. \textit{Grounding-DINO} processes the input image \( x_0 \) and garment description \( Y_0 \) using vision transformers and text embeddings to generate relevant bounding boxes. \textit{SAM} refines these boxes into a segmented mask \( M \in \{0, 1\}^{H \times W} \) (Figure \ref{fig:2}), we slightly extending \( M \) for smoother edges. This two-step approach ensures robust initial bounding boxes and accurate mask refinement, crucial for handling complex garment style. 

\subsubsection{Denoising-UNet}
\textit{Denoising-UNet} employs a latent diffusion model within the latent space of a variational autoencoder (VAE) comprising an encoder \( \mathcal{E} \) and a decoder \( \mathcal{D} \) \cite{VAE}. Starting with the latent representation of person image \( \mathcal{E}(x_0) \), noise is added through the diffusion model's forward process, resulting in \( z_T \). Using the mask \( M \) from \textit{Grounded-SAM}, the person image with the garment removed is represented as \( x_m = (1 - M) \odot x_0 \), where \( \odot \) denotes element-wise multiplication. Additionally, the input to \textit{Denoising-UNet} includes the latent representation of human densepose image \( p=\mathcal{E}(x_{\text{p}}) \), a garment texture image \( x_c \), and a textual description of the texture \( s \). The training loss function is formulated as:
\begin{equation}
\mathbb{E}_{z_T , t, M, p, \mathcal{E}(x_m), x_c , s, \epsilon \sim \mathcal{N}(0,I)} \left[ \| \epsilon - \epsilon_\theta (z'_T , t, x_c, s) \|_2 \right]
\end{equation}
where \( z'_T = [z_T, M, p, \mathcal{E}(x_m)] \). These latents are concatenated along the channel dimension, and the convolutional layers of the UNet are expanded to accommodate 13 channels, initialized with zero weights. 

To preserve the identity of the person and maintain the integrity of the unedited regions in the fashion image, we merge the edited fashion image \( x' \), generated by the decoder \( \mathcal{D} \) during the inference process, with the original model image \( x_0 \). The final composite image \( x_g \) is computed as:
\begin{equation}
x_g =  (1 - M) \odot x_0 + M \odot x',
\end{equation}

\subsection{Texture Injection and Refinement Mechanism}
To inject and preserve the intricate texture details in the generated garments, we propose a texture injection and refinement mechanism. This approach begins with a decoupled cross-attention mechanism that preliminarily aligns the textures of the input image with those of the generated output. Additionally, we introduce DP-UNet, specifically designed to further enhance and refine these texture details.

\subsubsection{Decoupled Cross-Attention Mechanism}
Inspired by the Image Prompt Adapter\cite{ipadapter}, we use a decoupled cross-attention mechanism for multimodal prompt control. Specifically, we decoupled the attention heads for text and image embeddings, allowing independent control over text and visual prompts. Let \( Q \) represent the query matrices derived from the main UNet's intermediate representation, while \( K \) and \( V \) denote the key and value matrices obtained from the text embeddings \( c_t \). The output of the cross-attention layer is given by:
\begin{equation}
\label{eq3}
\text{Attention}(Q, K, V) = \text{softmax}\left(\frac{Q K^\top}{\sqrt{d}}\right) V,
\end{equation}
where \( K = c_t W_k \) and \( V = c_t W_v \). Similarly, let \( K' \) and \( V' \) represent the key and value matrices derived from the image embeddings \( c_i \), with \( K' = c_i W'_k \) and \( V' = c_i W'_v \). Here, \( W_k \), \( W_v \), \( W'_k \), and \( W'_v \) are the weight matrices of the trainable linear projection layers. By adjusting the parameter \( \lambda \) during inference, the final formulation of the decoupled cross-attention mechanism is expressed as:
\begin{equation}
Z = \text{Attention}(Q, K, V) + \lambda \cdot \text{Attention}(Q, K', V'),
\end{equation}
When \( \lambda = 0 \), the model reverts to the original text-to-image diffusion model. We initialize the feature projection and cross-attention layers using IP-Adapter-Plus\footnotemark[1].
\footnotetext[1]{https://huggingface.co/h94/IP-Adapter}

\subsubsection{DP-UNet}
To preserve intricate garment textures and refine details, we introduce \textit{DP-UNet}, addressing the limitations of the original \textit{Denoising-UNet} in handling high-frequency details. For complex garment patterns, relying solely on CLIP to extract image features is insufficient.

Specifically, DP-UNet enhances these details by incorporating a refinement step that focuses on high-frequency features. Starting with the latent representation of the texture image \( \mathcal{E}(x_c) \), we first pass it through a frozen, pre-trained U-Net. During the downsampling process, the encoder of the pre-trained U-Net extracts detailed features \( f_c \) from the texture image. These features are subsequently concatenated with the corresponding features from the same layer of the denoising-UNet, facilitating the model's ability to accurately reconstruct the texture. Let \( Q \) represents the query matrix, \( K \) the key matrix, and \( V \) the value matrix. For texture feature \( f_i \) from the decoupled cross-attention layer and detail feature \( f_c \), we define:
\begin{equation}
Q = f_i W_q, \quad K = [f_i; f_c] W_k, \quad V = [f_i; f_c] W_v,
\end{equation}
The self-attention is computed on the combined features as Equation \ref{eq3}. Which \( W_q \), \( W_k \), and \( W_v \) as the weights of the self-attention layer in the denoising-UNet.
We use the DP-UNet from SDXL-Inpainting\footnotemark[2]. DP-UNet leverages the rich generative prior of the pre-trained text-to-image diffusion model, complementing the detailed features often overlooked by the decoupled cross-attention layer. This improvement allows DP-UNet to better handle complex garment patterns, ensuring the final images align closely with input descriptions and references. By incorporating this dedicated self-attention module, DP-UNet significantly enhances texture detail and overall image quality.
\footnotetext[2]{https://huggingface.co/diffusers/stable-diffusion-xl-1.0-inpainting-0.1}

\begin{figure*}[t]
    \centering
    \includegraphics[width=\textwidth]{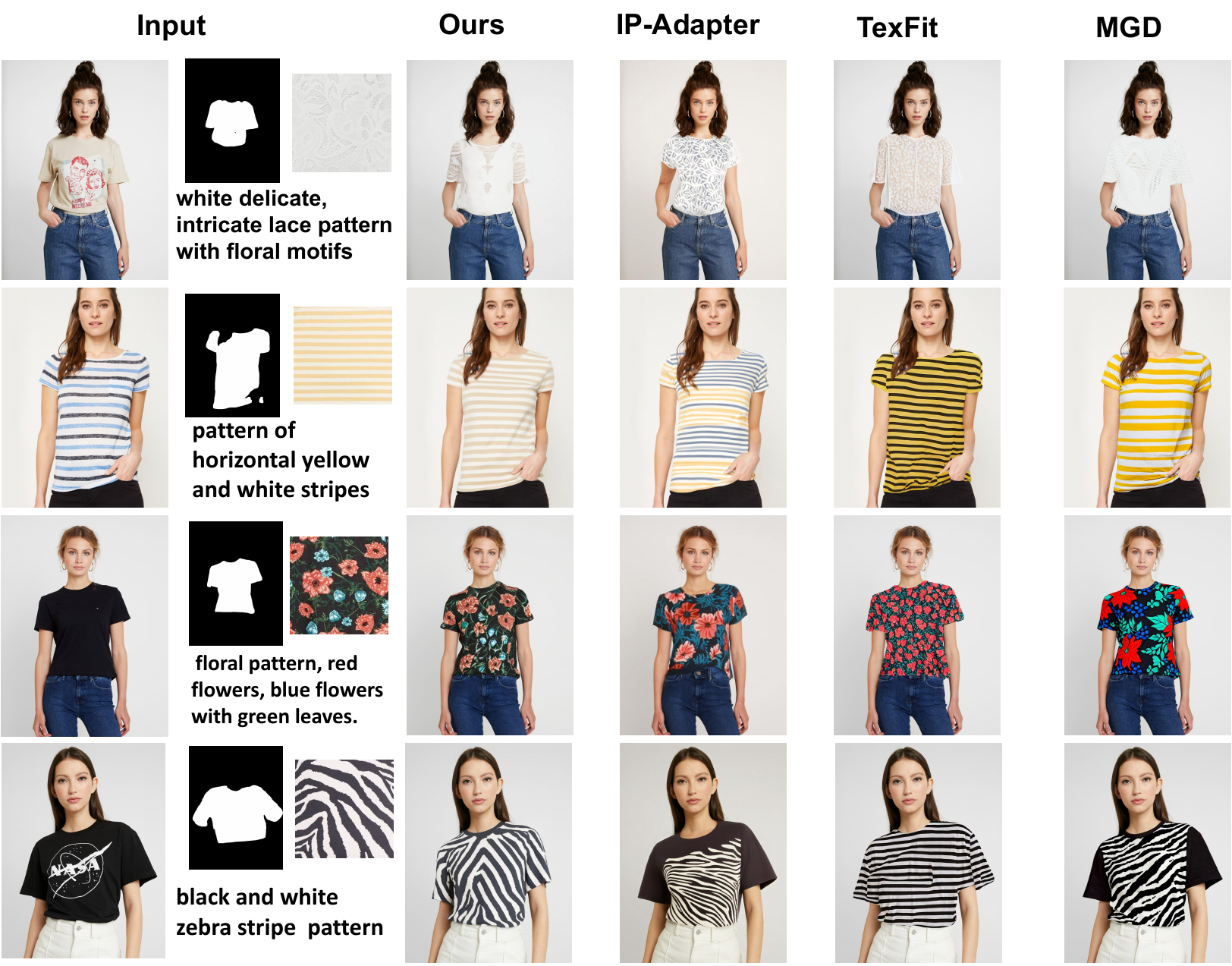}
    \caption{\small Qualitative comparison of images generated using our approach and baseline methods. The figure compares our method (Ours) with IP-Adapter, TexFit, and MGD across various garment textures and patterns.}
    \label{fig:5}
    \vspace{-0.5em}
\end{figure*}
\subsection{DPDEdit Datasets}
 The current fashion datasets lack the necessary multimodal information for the task we aim to address. To address this limitation, we extend the virtual try-on domain dataset VITON-HD to better align with our specific requirements. VITON-HD is a high-resolution dataset specifically designed for fashion applications, containing image pairs with a resolution of 1024 × 768 pixels. Each pair consists of a garment image and a corresponding model image wearing the garment. The dataset comprises 11,647 items for the training set and 2,032 items for the test set. For each garment \( C \) and its corresponding mask \( M_C \) in the dataset, we extract fabric textures using a sliding window of 128 × 128 pixels, selecting only patches \( X \) that are fully contained within the garment mask \( M_C \). To prevent resampling of specific regions of the garment, we use a stride of 64 pixels (128/2) in both horizontal and vertical directions. For garments with limited fabric area, where no suitable patch can be found within \( M_C \), we reduce the window size to 64 × 64 pixels to ensure at least one patch \( X \) can be extracted for each garment \( C \).  For images with multiple textures, we selected the primary texture located at the center of the garment.
 Then, we input the extracted fabric texture images into the multimodal large language model LLaVA to generate a textual description of the texture pattern image. We employ LLaVa v1.6-34b\footnotemark[3]
 \footnotetext[3]{https://huggingface.co/liuhaotian/llava-v1.6-34b} for this annotation task. Considering the distinctive features of fashion garment images, it’s crucial for the model to concentrate on key attributes like color, texture, fabric material, and pattern. To achieve accurate annotations for the texture images, we specifically highlighted these elements in our instructions. For more details of datasets construction, please refer to the supplementary materials.

\begin{table*}[h]
\centering
\caption{\small Quantitative results comparing the performance of our method with baseline methods across various modalities. Lower FID and LPIPS values indicate better image fidelity, while higher CLIP-I and CLIP-Score values reflect better alignment with textual descriptions and texture image.}
\label{tab:performance_comparison}
\begin{adjustbox}{max width=\textwidth}
\begin{tabular}{lcccccccc}
\toprule
\multirow{2}{*}{\textbf{Method}} & \multicolumn{4}{c}{\textbf{Modalities}} & \multicolumn{4}{c}{\textbf{Performance Metrics}} \\
\cmidrule(lr){2-5} \cmidrule(lr){6-9}
 & \textbf{Text} & \textbf{Mask} & \textbf{Pose} & \textbf{Texture} & \textbf{FID} $\downarrow$ & \textbf{LPIPS} $\downarrow$ & \textbf{CLIP-I} $\uparrow$ & \textbf{CLIP-Score} $\uparrow$ \\
\midrule
SD v1.5 inpaint & \checkmark & \checkmark & & & 18.62 & 0.331 & 0.435 & 25.26 \\
Texfit & \checkmark & \checkmark & & & 12.63 & 0.211 & 0.521 & \textbf{28.18} \\
MGD & \checkmark & \checkmark & \checkmark & & 11.87 & 0.243 & 0.459 & 25.38 \\
SDXL+ControlNet+IP-Adapter & \checkmark & \checkmark & \checkmark & \checkmark & 12.85 & 0.168 & 0.708 & 26.83 \\
\rowcolor{Gray}
DPDEdit (Ours) & \checkmark & \checkmark & \checkmark & \checkmark & \textbf{8.04} & \textbf{0.142} & \textbf{0.917} & 26.42 \\
\bottomrule
\end{tabular}
\end{adjustbox}
\end{table*}

\begin{table}[h]
\vspace{-1.0em}
\centering
\caption{Quantitative ablation study results for DPDEdit on the extended VITON-HD test dataset}
\label{tab:3}
\setlength{\tabcolsep}{3pt} 
\begin{tabular}{lcccc}
\toprule
\textbf{Module} & \textbf{FID} $\downarrow$ & \textbf{LPIPS} $\downarrow$ & \textbf{CLIP-I} $\uparrow$ & \textbf{CLIP-S} $\uparrow$ \\
\midrule
SDXL Inpainting & 14.54 & 0.308 & 0.515 & 27.13 \\
+Grounded-SAM & 12.49 & 0.224 & 0.534 & 27.85 \\
++DP-UNet & 9.17 & 0.165 & 0.774 & 26.61 \\
\bottomrule
\end{tabular}
\vspace{-1.0em}
\end{table}

\section{Experiments}
\subsection{Experimental Settings}

\noindent\textbf{Implementation Details}.
In our experiments, we employ the SDXL-inpainting model as the base model and use pre-trained IP-Adapter Plus weights to initialize our Decoupled Cross-Attention layer. Additionally, we utilize OpenCLIP ViT-H/14 as the image encoder. DPDEdit is trained using the extended VITON-HD dataset. We employ a two-stage training strategy. In the first stage, the DP-UNet component is excluded, allowing the primary focus to be on training the denoising-UNet and cross-attention layers. In the second stage, DP-UNet is introduced to enhance texture details. At this point, the denoising-UNet is frozen, and only the parameters within the cross-attention layers of the DP-UNet are updated. 


\subsection{Evaluation Metrics}
We use FID and LPIPS to evaluate image quality, CLIP-S to assess text-image alignment in the masked editing region, and CLIP-I to measure the consistency between the generated garment texture and the input texture image.

\subsection{Comparison to SOTA Methods}
We compare our DPDEdit with SOTA models, including Stable Diffusion inpainting pipeline, TexFit, MGD and IP-Adapter. For fairness, all methods (except MGD) are retrained on the extended VITON-HD dataset.
Table \ref{tab:performance_comparison} presents the quantitative results on the extended VITON-HD test dataset. The text-only conditioned method, TexFit, demonstrates competitive performance in FID (12.63) and LPIPS (0.211) metrics when compared to multimodal approaches MGD and IP-Adapter. This indicates that with accurate localization of the editing region, text-only editing method can also produce high-quality images. DPDEdit achieves the lowest FID (8.04) and LPIPS (0.142) scores, demonstrating its superior performance in generating high-fidelity fashion images.

However, DPDEdit shows slightly lower performance in CLIP-S compared to text-driven image editing methods. This difference can be attributed to the multimodal nature of our approach, which does not rely solely on text for image generation, leading to a less precise alignment with text descriptions. On the other hand, our method outperforms other comparison methods in CLIP-I, including IP-Adapter, which also utilizes texture images as a condition. This performance indicates that DPDEdit effectively captures and reproduces the fine details of the input texture images, ensuring a high degree of consistency in the generated fashion textures. We also present the qualitative comparison to evaluate our method. As we show in Figure \ref{fig:5}, while DPDEdit does not achieve the highest scores in the CLIP-S metric, it generally better than other competing methods both in visual realism and alignment with the texture image. This observation suggests a disparity between the garment textures conveyed through textual descriptions and those present in reality, underscoring the importance of integrating the texture image modality in our approach. In comparison to IP-Adapter, our approach achieves a higher degree of alignment with the input textures, demonstrating the effectiveness of the proposed texture injection and refinement mechanism. For more qualitative results, please refer to the supplementary materials.

\begin{figure}[t]
    \centering
    \includegraphics[width=\columnwidth]{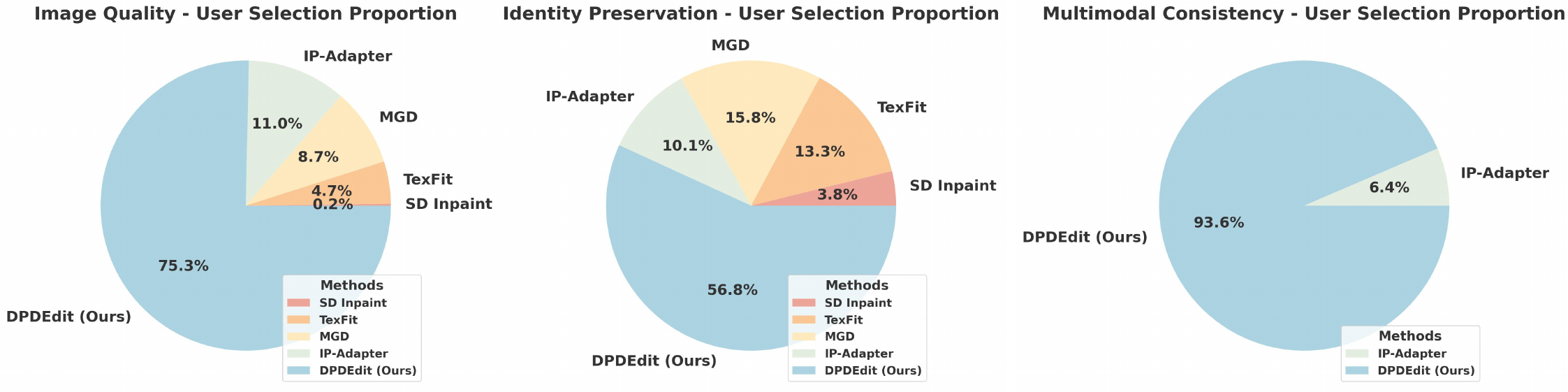}
    \caption{\small Results of the human feedback evaluation comparing our proposed method with baseline methods.}
    \label{fig:eva}
    \vspace{-1.0em}
\end{figure}

To ensure that our quantitative results align with human perspectives, we perform a human-subject study to evaluate our method through human judgment. We recruit 23 participants from design-related fields to evaluate 2,032 sets of result images from the test set. For each set, participants need to select the generated image that exhibits the best performance in terms of image quality, identity preservation and multimodal consistency. For the multimodal consistency metric, we only consider methods utilizing IP-Adapter with the same input modalities as our approach. The detailed results of the image selection are presented in Figure \ref{fig:eva}. Our method consistently outperforms the other methods across all evaluation criteria.

\begin{figure}[h]
    \vspace{-1.0em}
    \centering
    \includegraphics[width=\columnwidth]{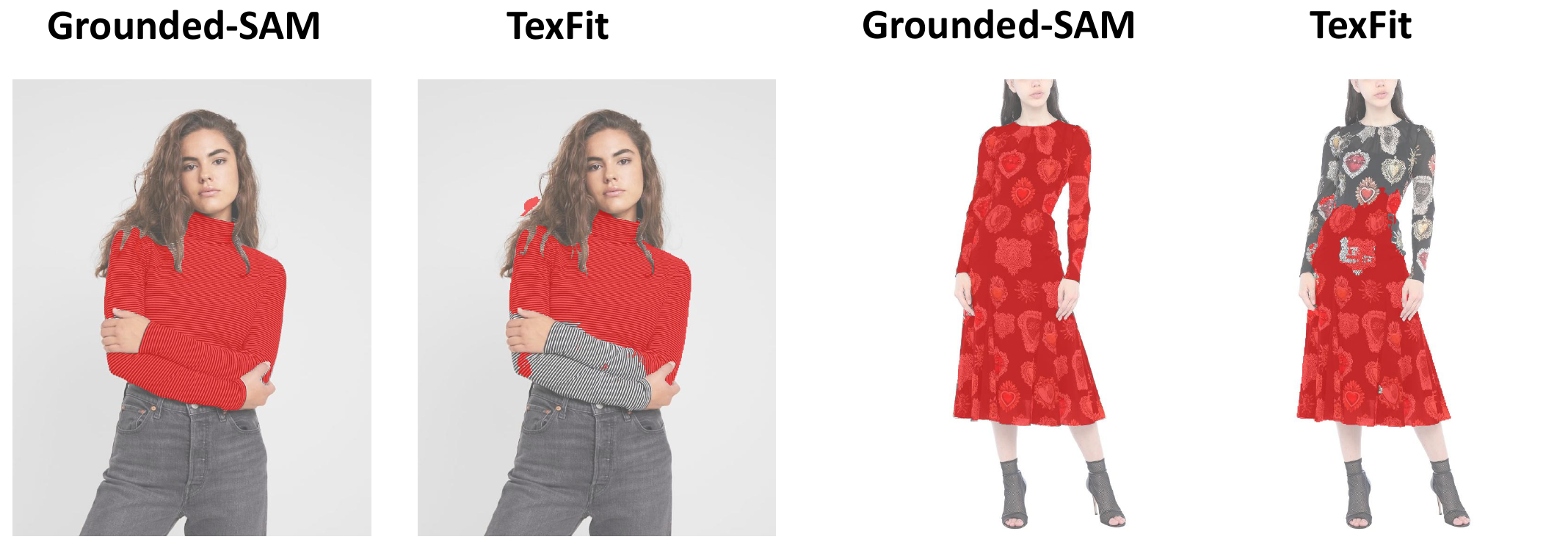}
    \caption{\small Comparison of editing region masks produced by TexFit and Grounded-SAM across different garment types.}
    \label{fig:6}
    \vspace{-1.0em}
\end{figure}

\subsection{Ablation Study}
\begin{figure}[h]
    \vspace{-0.5em}
    \centering
    \includegraphics[width=\columnwidth]{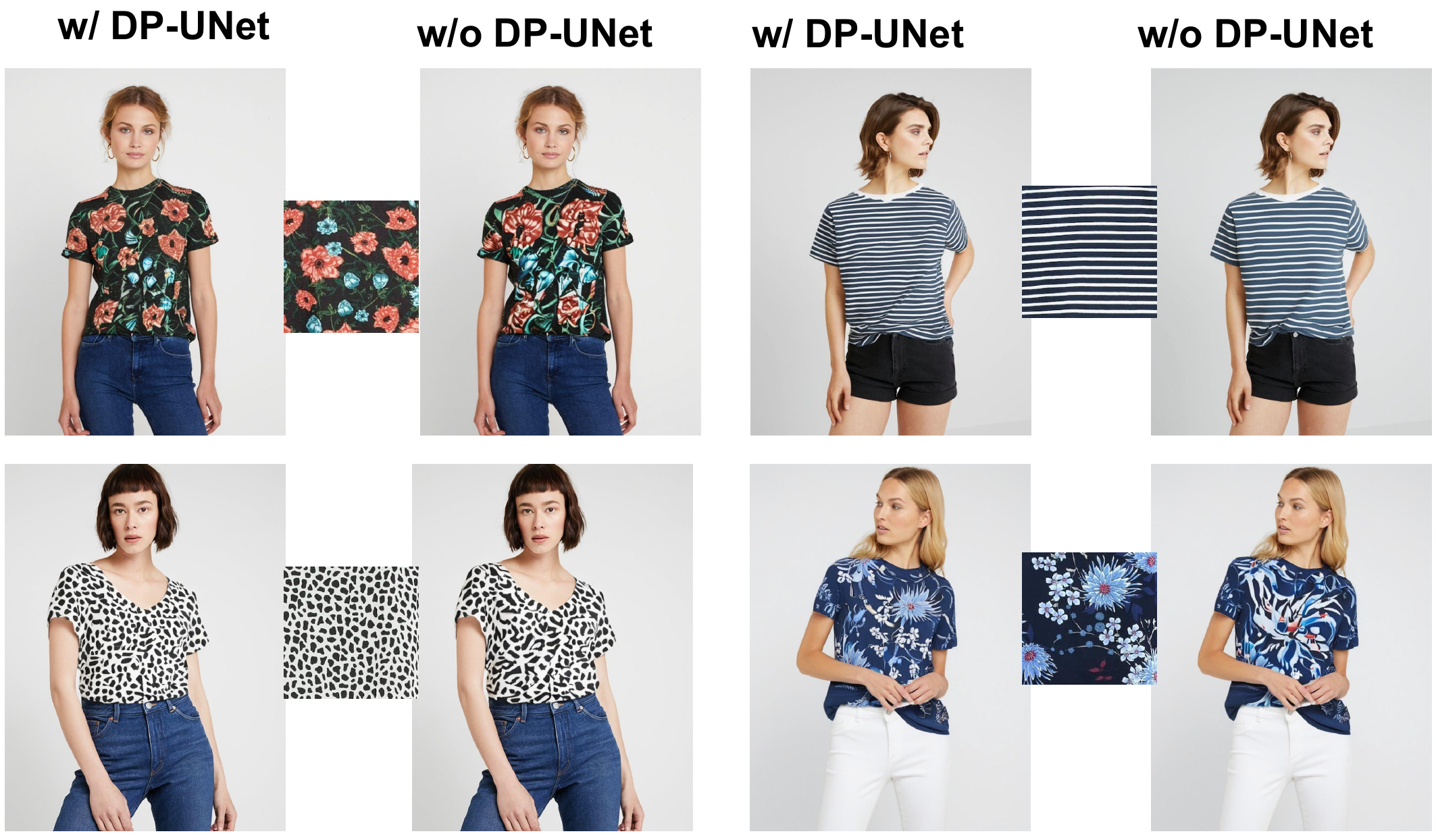}
    \caption{\small Ablation study on DP-UNet.}
    \label{fig:7}
    \vspace{-0.5em}
\end{figure}

We performed an ablation study on the Grounded-SAM and DP-UNet components of the proposed method to evaluate their effectiveness in localizing the garment editing regions and preserving the fine-grained details of garment textures. The qualitative results of Grounded-SAM are shown in Figure  \ref{fig:6}. Grounded-SAM exhibits greater accuracy in identifying editing regions compared to TexFit, especially in cases involving complex body poses and varied garment styles. The qualitative results of DP-UNet can be referenced Figure \ref{fig:7}. Images on the left in each pair are generated with DP-UNet, demonstrating improved pattern accuracy and consistency across different designs, while images on the right are without DP-UNet, showing less precise alignment.Furthermore, we conduct a quantitative evaluation on extended VITON-HD test dataset in Table \ref{tab:3}, we see that using  Grounded-SAM(replaces TexFit) and DP-UNet quantitatively improves Image fidelity and multimodal coherence, which is aligned with our qualitative results.

\section{Conclusion}
In this paper, we introduce DPDEdit, a novel method for fashion image editing guided by multimodal conditions. Our approach integrates textual descriptions, human poses, and garment textures to achieve localized editing in fashion images. DPDEdit utilize Grounded-SAM to ensures precise localization of garment regions. The proposed texture injection and refinement mechanism enables fine-grained control over the generated images. To address the challenges posed by this new task, we extend the existing VITON-HD dataset for training and evaluation purposes. Experimental results on this extended dataset demonstrate the superiority of our method, surpassing state-of-the-art techniques in terms of image fidelity and alignment with multimodal inputs.

\section*{Acknowledgment}
This work is partially supported by the National Natural Science Foundation of China (Grant No. 62176165) and the Stable Support Project for Shenzhen Higher Education Institutions (Grant No. 20220718110918001). 

This research is also supported by the GSFEI Top Scholar Award from the Graduate School of the University of Washington, and the Start-Up Funding awarded to Prof. Zhi-Qi Cheng.

\bibliographystyle{IEEEbib}
\bibliography{DPDEdit}
\end{document}


\title{DPDEdit: Detail-Preserved Diffusion Models for \\ Multimodal Fashion Image Editing}

\author{Anonymous ICME submission}

\maketitle
\appendix
\subsection{Experiments Details}
\subsubsection{Baselines}
We select four diffusion model-based image editing methods as our comparison baselines. For text-only inputs, we employ the Stable Diffusion inpainting pipeline and fashion image editing method TexFit\cite{texfit}, with the strength parameter of both methods adjusted to 0.9. For multimodal conditional inputs, MGD\cite{MGD} integrates text, human pose, sketch, and mask guidance through the text inversion technique. We substitute the text with a description of the texture pattern to be generated while keeping the other conditions unchanged. To ensure compatibility with our method's modality inputs, we utilize the Stable Diffusion XL model, integrated with ControlNet\cite{controlnet} for pose and IP-Adapter for texture images. The conditioning scale for ControlNet networks is set to 0.6, while the IP-Adapter\cite{ipadapter} scale is set to 0.5. For consistency, all methods (except MGD) are retrained on the extended VITON-HD dataset and inference on the same test set, with input masks generated by Grounded-SAM.

\subsubsection{Evaluation Metrics}
We utilize Fréchet Inception Distance (FID)\cite{FID} and Learned Perceptual Image Patch Similarity (LPIPS)\cite{LPIPS} to quantitatively evaluate the fidelity of the generated fashion images. Furthermore, to determine the alignment between the edited fashion images and the input text prompts, we employ the CLIP Score (CLIP-S)\cite{clipscore}. We calculate CLIP-S by focusing only on the masked editing region of the fashion image. To evaluate how closely the generated garment matches the input fabric texture, we crop a 128 × 128 pixel portion of the image to capture the texture of the generated garment and compute the CLIP score between the cropped region and the input texture image, denoted as CLIP-I.

\subsubsection{Training and Inference Details}
DPDEdit is trained on the extended VITON-HD\cite{viton-hd} dataset, which consists of 11,647 texture image-text pairs. For data augmentation\cite{stableviton}, we apply horizontal flipping with a probability of 0.5 and random affine transformations, including shifting and scaling (limited to 0.2, with a probability of 0.5) to the multimodal inputs. The model is trained on a single machine equipped with 8 A6000 GPUs for 65k steps, with a batch size of 8 per GPU. We employ the AdamW\cite{adam} optimizer with a fixed learning rate of 1e-5 and a weight decay of 0.01. To facilitate classifier-free guidance\cite{cfg}, we use a probability of 0.05 to drop either the text or the texture image individually, and a probability of 0.05 to drop both simultaneously. During inference, we utilize the DDIM\cite{ddim} sampler with 30 steps, setting the guidance scale to 5.0, which has been found effective in practice. When only the texture image prompt is used, the text prompt is left empty, and \( \lambda \) is set to 1.0. Additionally, a batch size of 2 is used during inference to efficiently manage GPU memory. To ensure reproducibility across different inference runs, we use a random seed of 42.

\begin{figure}[h]
    \centering
    \includegraphics[width=\columnwidth]{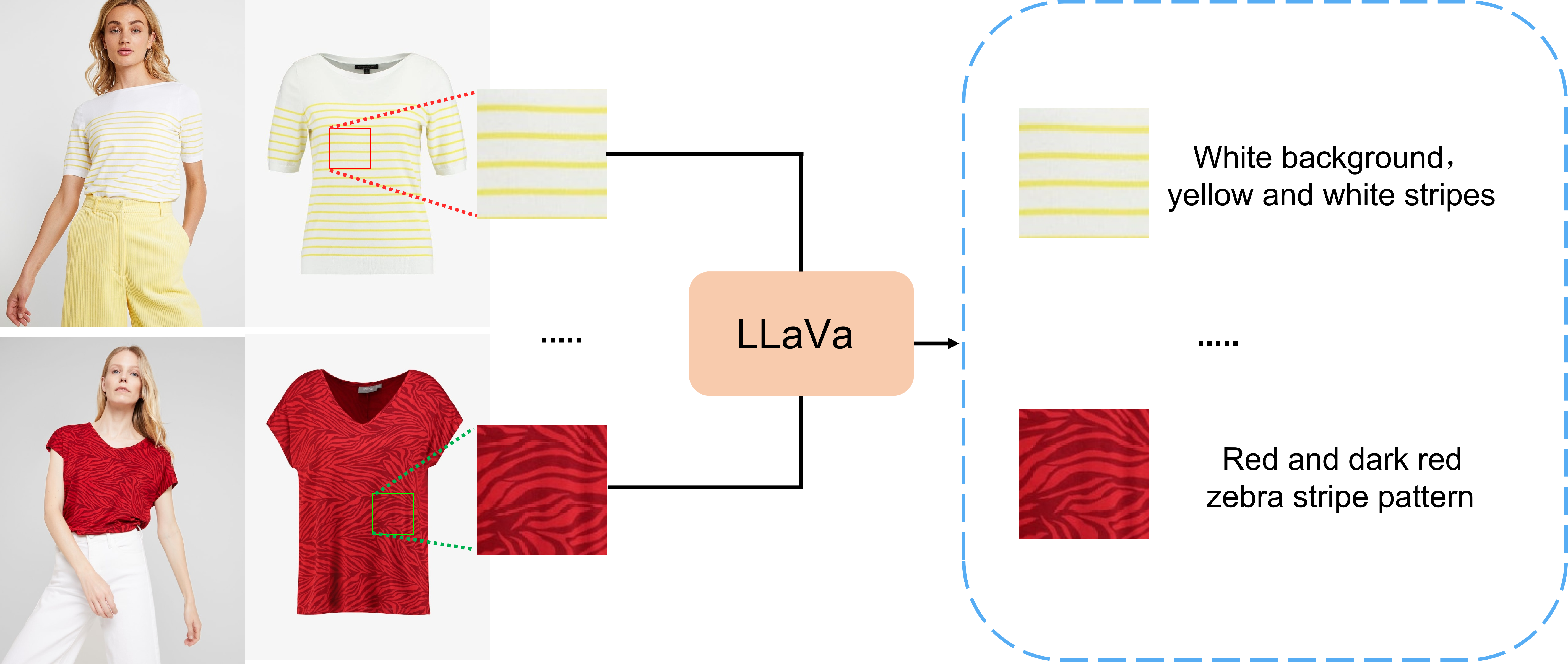}
    \caption{Illustration of extending the VITON-HD dataset to generate paired texture images and textual descriptions.}
    \label{fig:4}
\end{figure}

\subsection{Datasets Construction}
\begin{figure*}[t]
    \centering
    \includegraphics[width=\linewidth]{IEEE Conference Template - ICME 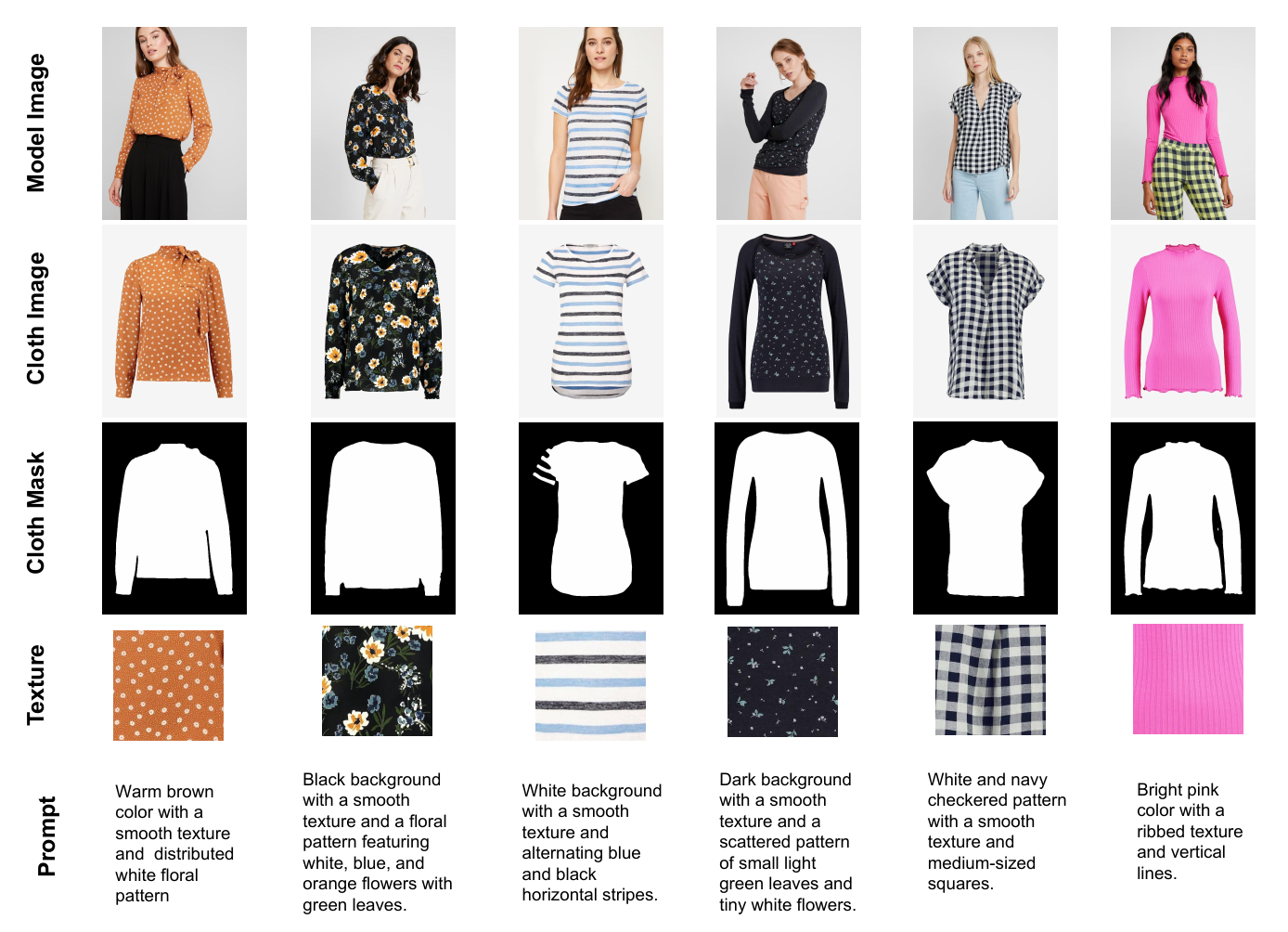}
    \caption{Samples from the extended VITON-HD dataset, illustrating a diverse array of garment textures paired with detailed text descriptions.}
    \label{fig:dataset}
\end{figure*}

To create a paired dataset of garment texture images and text descriptions, we utilized LLaVA1.6-34B\cite{llava} to annotate the fashion texture images (Figure \ref{fig:4}). Due to the low resolution of the texture images extracted from garments, we upscaled the garment texture to 256x256 to display more detailed patterns. To diversify the model's responses, we employed various types of instructions during the dialogue. Considering the distinctive features of fashion garment images, it's crucial for the model to concentrate on key attributes like color, texture, fabric material, and pattern. To achieve accurate annotations for the texture images, we specifically highlighted these elements in our instructions.The instructions as shown in Table \ref{tab:instructions}. The dataset generated using this strategy is shown in Figure \ref{fig:dataset}. This method facilitates the creation of a diverse set of garment textures paired with detailed text descriptions, providing robust support for our task.

To assess the quality of the dataset annotations, we randomly sampled 1,000 images for manual review. Five volunteers evaluated the texture image descriptions by scoring them. Each volunteer rated a total of 200 descriptions on a scale from 0 to 5. The evaluation criteria included the consistency of the description with the texture image in terms of color, texture, pattern, and material.The results are shown in Table~\ref{tab:annotation_evaluation}.
\begin{table}[h]
\centering
\caption{Manual Evaluation of Annotation Quality(Scale
from 0 to 5.)}
\label{tab:annotation_evaluation}
\begin{tabular}{lccccc|c}
\toprule
\textbf{Volunteer} & \textbf{V1} & \textbf{V2} & \textbf{V3} & \textbf{V4} & \textbf{V5} & \textbf{Average} \\
\midrule
\textbf{Score}     & 4.23       & 3.86       & 4.15       & 4.11       & 4.57       & 4.184 \\
\bottomrule
\end{tabular}
\end{table}
The average score from human evaluation indicates that our annotation strategy effectively enables the model to focus on the key information in texture images.

\subsection{Computational Cost}
To assess the computational cost of our approach, we conducted a comparative evaluation between different methods. The results are summarized in Table~\ref{tab:computational_cost}.

\begin{table}[h]
\centering
\caption{Computational Cost Comparison Across Methods}
\label{tab:computational_cost}
\begin{tabular}{lccc}
\toprule
\textbf{Method} & \textbf{Params (M)} & \textbf{Mem (G)} & \textbf{Time (×1.0)} \\
\midrule
SDXL & 3468 & 6.36 & ×1.0 \\
SDXL + ControlNet & 3892 & 12.67 & ×1.32 \\
SDXL + ControlNet + IP-A & 3914 & 14.21 & ×1.54 \\
\textbf{Ours} & \textbf{5021} & \textbf{16.37} & \textbf{×1.69} \\
\bottomrule
\end{tabular}
\end{table}

Here, \textbf{Params} denotes the number of trainable parameters (in millions), \textbf{Mem} indicates the inference memory consumption (in gigabytes), and \textbf{Time} reflects the relative inference duration normalized to the baseline SDXL method.

As seen in the results, our model requires a slight increase in computational resources. However, this increase is within an acceptable range, especially considering the substantial improvements in result quality that our method achieves. The trade-off between computational overhead and output quality underscores the efficiency and effectiveness of our proposed approach.
\begin{figure*}[t]
    \centering
    \includegraphics[width=0.92\textwidth]{IEEE Conference Template - ICME 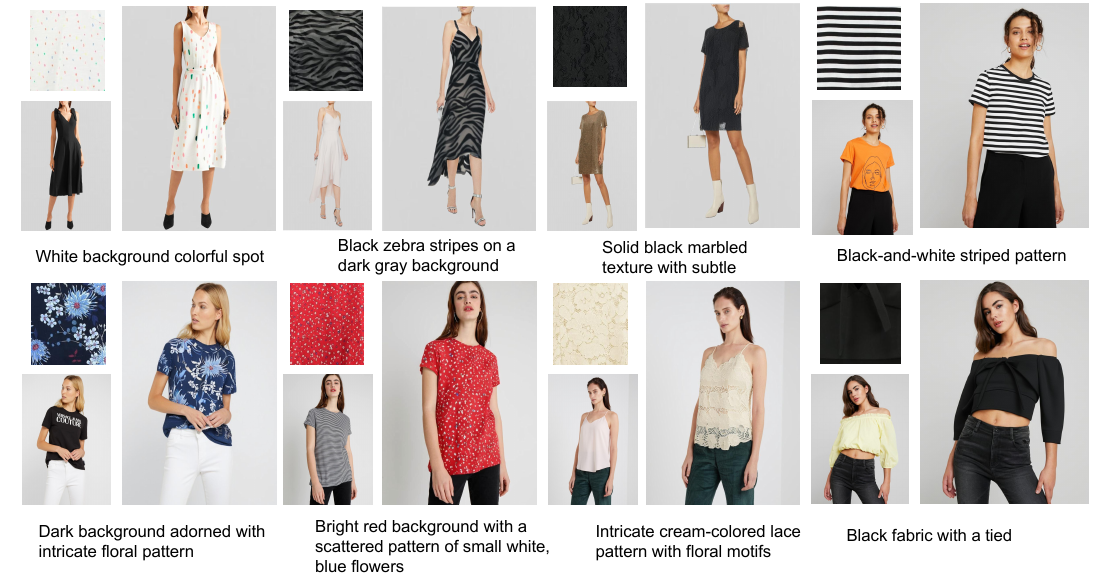}
    \caption{\small Sample images generated using the proposed DPDEdit method. For each sample, we show the input image(bottom left), fabric texture(top left), descriptive caption of the texture image(bottom of the sample), and the final generated result}
    \label{fig:1}
\end{figure*}
\subsection{Additional Qualitative Results of DPDEdit}
\begin{figure*}
    \centering
    \includegraphics[width=\linewidth]{IEEE Conference Template - ICME 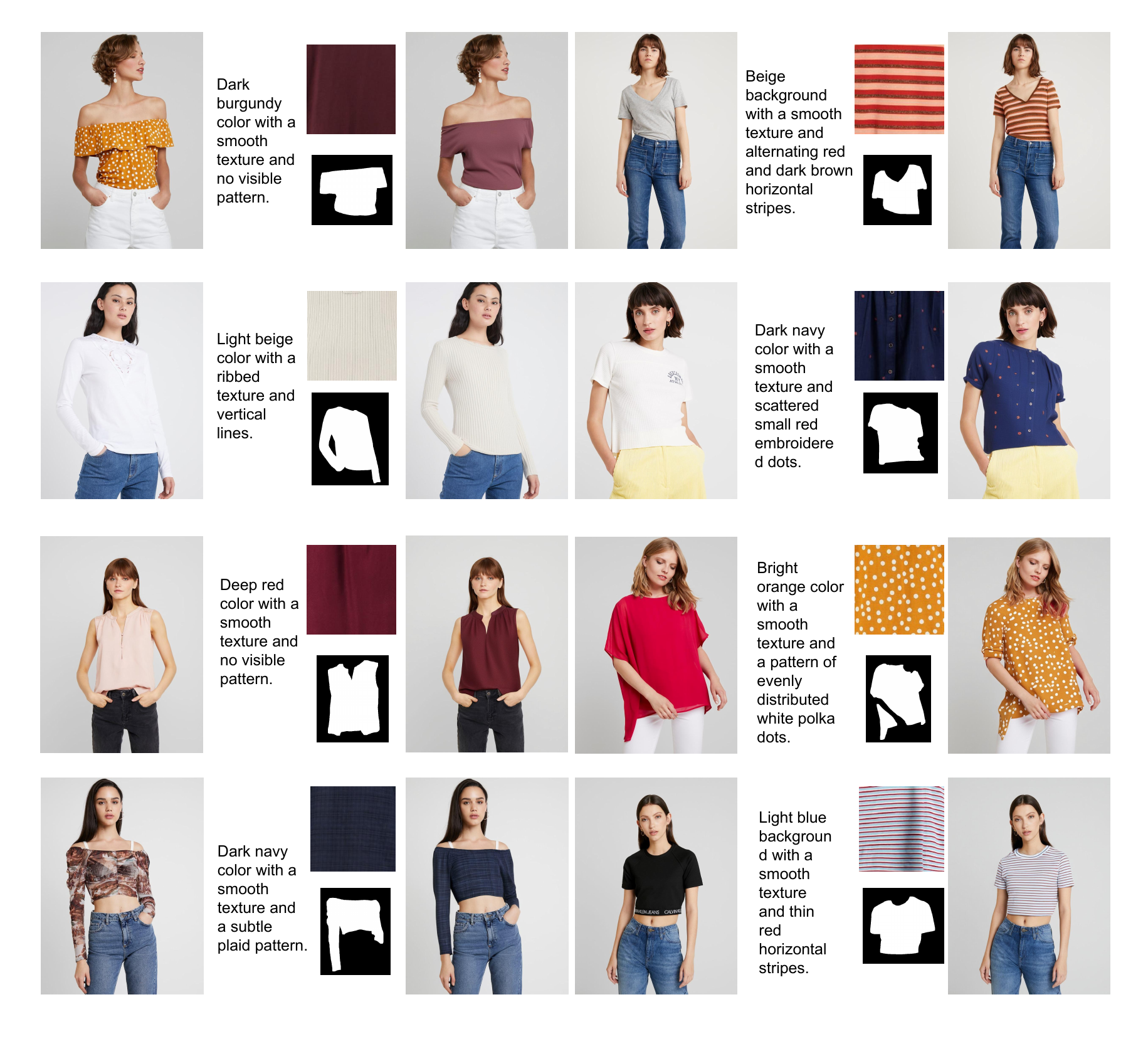}
    \caption{Qualitative results on the extended VITON-HD test set.}
    \label{fig:show_1}
\end{figure*}

\begin{figure*}
    \centering
    \includegraphics[width=\linewidth]{IEEE Conference Template - ICME 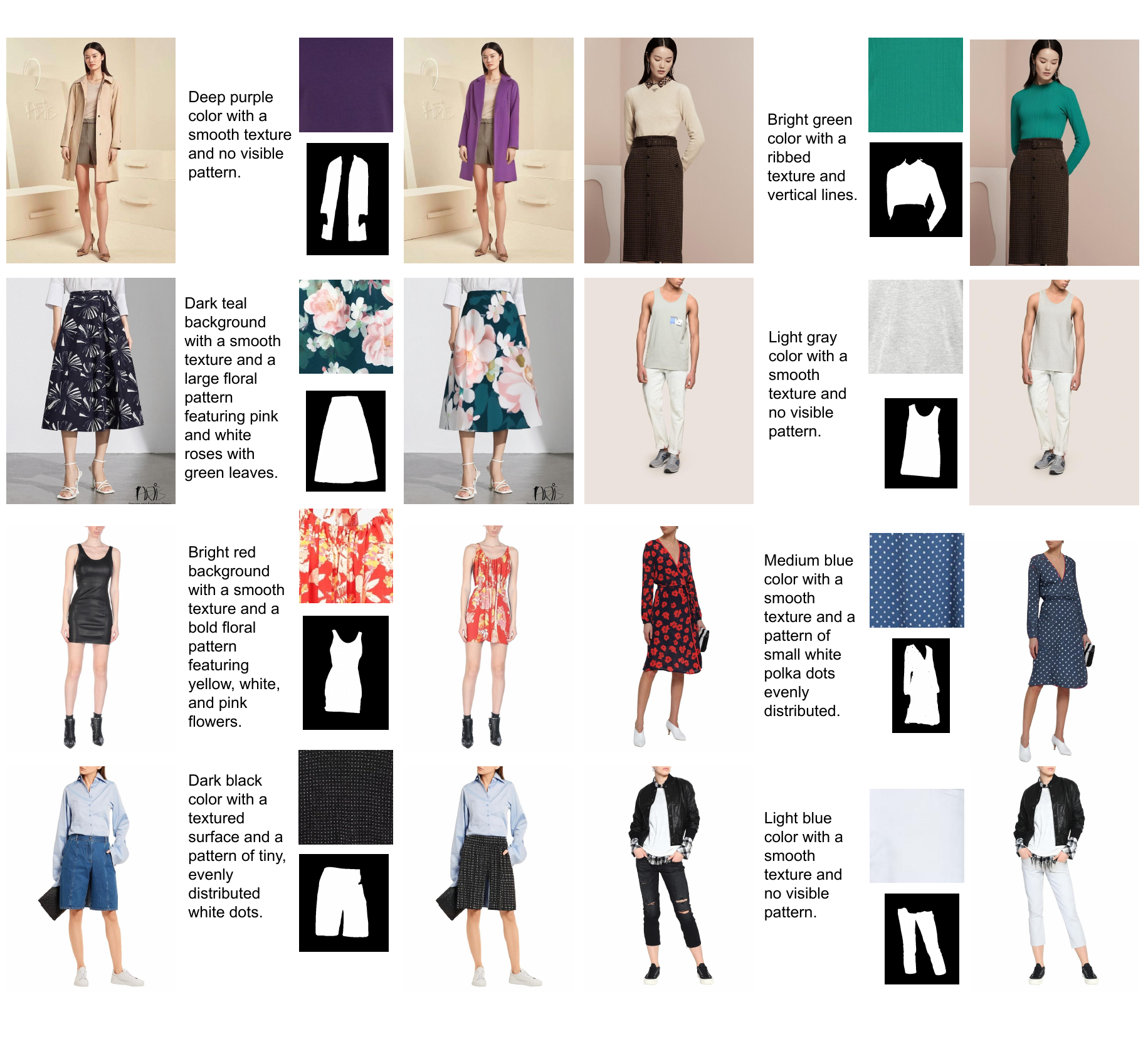}
    \caption{Qualitative results of DPDEdit on a broader range of datasets, including fashion images from open-world scenarios and the Dresscode dataset.}
    \label{fig:show_2}
\end{figure*}

In this section, we present supplementary qualitative results to further demonstrate the effectiveness of DPDEdit. Figure \ref{fig:1} presents the generation results of our proposed DPDEdit across different clothing styles. Figure \ref{fig:show_1} showcases results on the extended VITON-HD test set, where the use of precise editing region masks generated by Grounded-SAM\cite{groundedsam} enables DPDEdit to seamlessly modify the color, texture, and patterns of target garments while maintaining the original design. Additionally, Figure \ref{fig:show_2} illustrates DPDEdit's performance on a broader range of datasets, including fashion images from open-world scenarios and other datasets such as Dresscode\cite{dresscode}. These results highlight DPDEdit's ability to edit fashion garments across various backgrounds and human poses, as well as its effectiveness in modifying different parts of garments, including the lower body and dresses.

\begin{table*}
\centering
\renewcommand{\arraystretch}{1.5} 
\caption{Instructions for Fashion Garment Image Annotation}
\label{tab:instructions}
\begin{tabular}{|l|}
\hline
1. \textit{You are a fashion designer,} describe the key features of this garment, focusing on its color, texture, fabric material, and pattern. \\ \hline
2. \textit{Identify the primary colors and textures} present in this garment image. \\ \hline
3. \textit{Describe the fabric material} of the garment in this image. What kind of texture does it exhibit? \\ \hline
4. \textit{You are tasked with designing a similar garment,} describe the color, texture, and pattern you observe in this image. \\ \hline
5. \textit{What are the standout features} of the garment’s texture and pattern in this image? \\ \hline
6. \textit{Provide a comprehensive analysis} of the garment's color, fabric material, and texture. \\ \hline
7. \textit{Describe the overall aesthetic} of the garment, focusing on the fabric's texture and pattern. \\ \hline
\end{tabular}
\end{table*}

\bibliographystyle{IEEEbib}
\bibliography{DPDEdit}